\documentclass{article}

\usepackage{arxiv}

\usepackage[utf8]{inputenc} 
\usepackage[T1]{fontenc}    
\usepackage{hyperref}       
\usepackage{url}            
\usepackage{booktabs}       
\usepackage{amsfonts}       
\usepackage{nicefrac}       
\usepackage{microtype}      
\usepackage{lipsum}		
\usepackage{graphicx}
\usepackage{natbib}
\usepackage{doi}

\usepackage{caption}
\usepackage{diagbox}

\usepackage{amsmath}
\usepackage{algorithmic}
\usepackage{stfloats}

\usepackage[table,x11names]{xcolor}
\usepackage{collcell}
\usepackage{array}
\usepackage{tikz}
\usepackage{pgfkeys}

\pgfkeys{/heat/.is family, /heat,
  Max colour/.initial = Green4,
  Min colour/.initial = Red1,
  max colour/.initial = SpringGreen3,
  mid colour/.initial = Yellow1,
  min colour/.initial = Red1,
  text colour/.initial = black,
  Min color/.style = {Min colour=#1},
  Max color/.style = {Max colour=#1},
  min color/.style = {min colour=#1},
  mid color/.style = {mid colour=#1},
  max color/.style = {max colour=#1},
  text color/.style = {text colour=#1},
  min/.initial = -1,
  mid/.initial = 0,
  max/.initial = 1,
  slider/.code={%
     \tikz{\shade[left color=\HVal{min colour},%
                  right color=\HVal{max colour}]%
        (current page.south west) rectangle ++(#1,12pt);
     }%
  }%
}
\newcommand\Heatset[1]{\pgfkeys{/heat, #1}}
\newcommand\HVal[1]{\pgfkeysvalueof{/heat/#1}}

\newcolumntype{H}{>{\collectcell\Heat}c<{\endcollectcell}}
\newcommand\Heat[1]{
\if\relax\detokenize{#1}\relax
\else%
  \pgfmathparse{int(100*(#1-\HVal{mid})/(\HVal{max}-\HVal{mid}))}
    \ifnum\pgfmathresult>100
        \edef\HeatCell{\noexpand\cellcolor{\HVal{Max colour}}}%
        \else\ifnum\pgfmathresult>0
            \edef\HeatCell{\noexpand\cellcolor{\HVal{max colour}!\pgfmathresult!\HVal{mid colour}}}%
            \else
                \pgfmathparse{int(100*(#1-\HVal{min})/(\HVal{mid}-\HVal{min}))}
                \ifnum\pgfmathresult>0  
                    \edef\HeatCell{\noexpand\cellcolor{\HVal{mid colour}!\pgfmathresult!\HVal{min colour}}}%
                    \else
                        \edef\HeatCell{\noexpand\cellcolor{\HVal{Min colour}}}
                    \fi
            \fi%
        \fi%
    \HeatCell\textcolor{\HVal{text colour}}{$#1$}%
    \fi%
}

\title{Deep Recursive Embedding for High-Dimensional Data}

\author{ 
    Zixia~Zhou
    \thanks{Z. Zhou is with the Department of Electronic Engineering, Fudan University, Shanghai 200433, China. E-mail: 16110720022@fudan.edu.cn.} \\
	Department of Electronic Engineering\\
	Fudan University\\
	\texttt{16110720022@fudan.edu.cn} \\
	\And
    Xinrui~Zu
	\thanks{X. Zu is with the faculty of Electrical Engineering, Mathematics and Computer Science (EEMCS), University of Twente, Drienerlolaan 5, 7522 NB Enschede, the Netherlands. E-mail: x.zu@student.utwente.nl.} \\
	Faculty of Electrical Engineering, \\Mathematics and Computer Science\\
	University of Twente\\
	\texttt{zuxinrui95@gmail.com} \\
	\And
    Yuanyuan~Wang
	\thanks{Y. Wang is with the Department of Electronic Engineering, Fudan University, Shanghai 200433, China and Key Laboratory of Medical Imaging Computing and Computer Assisted Intervention of Shanghai, Shanghai 200032, China. E-mail: yywang@fudan.edu.cn.} \\
	Department of Electronic Engineering\\
	Fudan University\\
	\texttt{yywang@fudan.edu.cn} \\
	\And
    Boudewijn~P.F.~Lelieveldt
	\thanks{Boudewijn P.F. Lelieveldt is with the Division of Image Processing, Department of Radiology, Leiden University Medical Center, Albinusdreef 2, 2333 ZA Leiden, the Netherlands. E-mail: B.P.F.Lelieveldt@lumc.nl.} \\
	Department of Radiology\\
	Leiden University Medical Center\\
	\texttt{B.P.F.Lelieveldt@lumc.nl} \\
	\And
	Qian~Tao
	\thanks{Qian Tao is with the Department of Imaging Physics, Delft University of Technology, Lorentzweg 1, 2628 CJ Delft, the Netherlands. E-mail: q.tao@tudelft.nl.} \\
	Department of Imaging Physics\\
	Delft University of Technology\\
	\texttt{q.tao@tudelft.nl} \\
}

\renewcommand{\shorttitle}{Deep Recursive Embedding for High-Dimensional Data}

\hypersetup{
pdftitle={A template for the arxiv style},
pdfsubject={q-bio.NC, q-bio.QM},
pdfauthor={Xinrui~Zu},
pdfkeywords={t-distributed stochastic neighbor embedding, uniform manifold approximation and projection, deep embedding network, deep recursive embedding, unsupervised learning},
}

\begin{document}
\maketitle

\begin{abstract}
	Embedding high-dimensional data onto a low-dimensional manifold is of both theoretical and practical value. In this paper, we propose to combine deep neural networks (DNN) with mathematics-guided embedding rules for high-dimensional data embedding. We introduce a generic deep embedding network (DEN) framework, which is able to learn a parametric mapping from high-dimensional space to low-dimensional space, guided by well-established objectives such as Kullback-Leibler (KL) divergence minimization. We further propose a recursive strategy, called deep recursive embedding (DRE), to make use of the latent data representations for boosted embedding performance. We exemplify the flexibility of DRE by different architectures and loss functions, and benchmarked our method against the two most popular embedding methods, namely, t-distributed stochastic neighbor embedding (t-SNE) and uniform manifold approximation and projection (UMAP). The proposed DRE method can map out-of-sample data and scale to extremely large datasets. Experiments on a range of public datasets demonstrated improved embedding performance in terms of local and global structure preservation, compared with other state-of-the-art embedding methods.
\end{abstract}


\section{Introduction}
Embedding high-dimensional data onto a low-dimensional manifold is of both theoretical and practical value. It can be used for a variety of applications such as data visualization, representation learning, unsupervised clustering, and data exploration \citep{Wang2017, bengio2013deep, bengio2014representation, Liu2017}. t-distributed stochastic neighbor embedding (t-SNE) \citep{JMLR:v9:vandermaaten08a} is among the most well-known and widely-used methods for high-dimensional data visualization, which preserves local similarity of high-dimensional data in a drastically reduced low-dimensional map (typically 2). t-SNE has found widespread applications in life science research among others in the last decade, and established itself as an important visualization tool in the scientific community \citep{Kobak453449}. Nonetheless, it is so far largely recognized as a visualization tool, as some practical issues may have limited its wider use in learning-based tasks. First, t-SNE does not immediately allow out-of-sample projection. Second, the computation of t-SNE is memory- and time-consuming, not well scalable to the extremely large datasets of today. Third, t-SNE (especially its fast implementations) focuses on local neighborhood, often losing global data structure of data.

A number of follow-up work have been proposed after the initial paper of t-SNE. For instance, Barnes-Hut-SNE (BH-SNE) \citep{JMLR:v15:vandermaaten14a}, A-tSNE \citep{Pezzotti2017} and FIt-SNE \citep{Linderman2019} were proposed to accelerate the computation and reduce memory usage of t-SNE. Recently, another DR method called uniform manifold approximation and projection (UMAP) \citep{McInnes2018} was introduced and became another popular visualization tool. UMAP has demonstrated competitive visualization performance as t-SNE, with notably reduced runtime and arguably better preservation of global structure \citep{McInnes2018}.

In the meanwhile, DR has also been studied in the deep learning field \citep{Hinton504, makhzani2015adversarial}. For example, the auto-encoder (AE) is a classical DR method \citep{Hinton504}, which embeds high-dimensional data through a DNN with an encoder-decoder architecture. AE can be trained with mean square error (MSE) loss and standard optimization, and the intermediate tensors at the information bottleneck are seen as low-dimensional representations of the high-dimensional data. Although its embedding performance is generally inferior to that of t-SNE or UMAP, AE suggests great potential of DNN for DR through unsupervised learning. 

We posit that the combination of DNN and dedicated, mathematics-guided embedding rules may further enrich the possibilities of DR and improve its performance. Embedding rules such as cross entropy and Kullback-Leibler (KL) divergence are more specific to the purpose of DR, compared to MSE which is intended for reconstructing data. Generally speaking, there exist two methodologies to combine DNN and DR. Intuitively, we may teach the DNN to learn from a referenced DR method. The recent deep learning multidimensional projection method \citep{Espadoto2020} follows this methodology, which first computes a t-SNE or UMAP map in conventional ways, then let the DNN learn the t-SNE or UMAP results by MSE. This method can achieve scalability and out-of-sample support, but its embedding performance has an upper limitation equal to that of the referenced method. The second methodology is to train a DNN with dedicated loss functions to directly capture the mathematical principles of embedding, therefore \emph{mathematics-guided}. The parametric t-SNE (ptSNE) proposed in 2009 \citep{pmlr-v5-maaten09a} is an early work in this direction, which employed the t-SNE loss function to train a restricted Boltzmann machine (RBM). ptSNE addresses the scalability issue, but still has some practical limitations. ptSNE involves a pre-training step and a fine-tuning step, which are complicated while not guaranteed to converge. Even with ideal convergence, ptSNE cannot exceed the performance of the original t-SNE. 

We seek to develop a parametric (i.e. able to embed new data points) and scalable (i.e. able to embed infinite data points) DR method that can find an intrinsic low-dimensional representation of data. In this paper, we propose a generic \emph{deep embedding network (DEN)} framework that is based on well-established embedding rules and optimized by modern DNN. In addition, to break the upper limit of the performance in classical DR methods, we propose a recursive training strategy called \emph{deep recursive embedding (DRE)}, to make use of latent representations to boost the DR performance further. 

Specifically, as we will show in later sections, the advantages of the proposed DRE method are four-fold: (1) It can readily map  out-of-sample data points. (2) It is scalable and memory-efficient. (3) It is flexible in the design of its architecture, loss, and training strategy. (4) It breaks the upper limit of t-SNE or UMAP.

\section{Related Work}
\subsection{Classical Dimensionality Reduction Methods}
Dimensionality reduction is a classical problem in machine learning. The most well-known linear DR method is principal component analysis (PCA) \citep{Jolliffe2014}, which computes low-dimensional representation by linearly projecting high-dimensional data points onto the first few principal components that preserves the largest variance. Other classical DR methods include multidimensional scaling (MDS) \citep{Buja2008}, Isomap \citep{balasubramanian2002isomap}, locally linear embedding (LLE) \citep{Roweis2000}, and stochastic neighbor embedding (SNE) \citep{hinton2002stochastic}. They usually produce better visualization results on nonlinear data compared with PCA, but with limitations in terms of scalability, speed, and stochasticity. In 2008, the t-SNE method \citep{JMLR:v9:vandermaaten08a}, a variant of SNE, was proposed, which later became one of the most popular methods in the research community for high-dimensional data visualization. t-SNE emphasizes local neighborhood in data with a modified assumption (t-distribution in low-dimensional space) over SNE, and can produce higher quality embedding results compared with other nonlinear manifold learning methods. 

Many developments followed up the original t-SNE \citep{Chan2018, Pezzotti2020}. An important work is Barnes-Hut t-SNE (BH t-SNE) \citep{JMLR:v15:vandermaaten14a}, which approximates the high-dimensional space with sparse distributions and uses the Barnes-Hut algorithm to speed up computation. The approximation largely reduces the algorithm complexity and memory consumption, enabling t-SNE to embed very large datasets that were previously impossible. A hierarchical stochastic neighbor embedding (HSNE) method was proposed to interactively render the visualization at different scales \citep{Pezzotti2016}. The hierarchical way of visualization allows small memory footprint by focusing on landmarks instead of all data. The authors also proposed an A-tSNE method \citep{Pezzotti2017}, which trades off speed and accuracy, to enable interactive data exploration. More recently, Linderman et al. proposed a fast Fourier transform-accelerated interpolation-based t-SNE (FIt-SNE) \citep{Linderman2019} to accelerate the implementation of t-SNE. In FIt-SNE, an effective late exaggeration strategy was also proposed for better clustering. Lately, another important embedding method UMAP was introduced \citep{McInnes2018}, which is rooted in mathematical foundations of Riemannian geometry and algebraic topology. UMAP is considered competitive with t-SNE in visualization, while faster and better scalable in implementation. It is also able to map out-of-sample data by graph embedding. However, it was reported that UMAP can be sensitive to the choice of hyper-parameters \citep{Espadoto2020}. Although some work argued that the global data structure can be better preserved by UMAP, other studies suggested that UMAP preserves the global structure in a way similar to t-SNE with the same initialization \citep{Kobak2019}. 

\subsection{Deep Learning Methods}
AE is a classical DNN-based DR method, which maps the data to a low-dimensional representation with an encoder. Through unsupervised learning, the encoder can generate a nonlinear embedding in the low-dimensional space. While AE uses the MSE loss, the ptSNE method \citep{pmlr-v5-maaten09a}, proposed by van der Maaten in 2009, adopted the same KL loss function as t-SNE. ptSNE consists of three separate steps: training RBMs, stacking the RBMs to construct a pre-trained NN, and finetuning the pretrained NN. Such a training process is cumbersome, and its performance is frequently poorer than that of the standard t-SNE due to the difficulty of optimization. Another DNN method was later proposed to directly cluster data, using a loss function similar to that of t-SNE, but based on estimated centroids \citep{xie2015unsupervised}. Very recently, a DL Projection method \citep{Espadoto2020} was introduced to learn from any referenced DR methods: it first selects a reference method (e.g. t-SNE, UMAP) to generate a projection of a subset; then trains a DNN to learn the embedding. The method is effective and intuitive, but its performance is restricted by its teachers. Although the network itself does not demand complex parameter settings, the front-end process (i.e. the DR method to generate the ground truth for supervised learning) still requires elaborate tuning. 

We believe there is still much potential in DNN for DR, given its strong capability of representation learning. In this work, we propose a dedicated DNN-based DR framework, named \emph{deep embedding network (DEN)}, to embed high-dimensional data in a unsupervised manner by well-established mathematical principles. Within this framework, we further propose a novel \emph{deep recursive embedding (DRE)} method, which can make use of the latent representations of the original data for further boosted DR performance. To demonstrate the flexibility of our framework, we introduce a two-stage \emph{t-SNE-UMAP loss}, which is able to combine the favorable properties of t-SNE and UMAP. We provide a further detailed comparison between our methods and other deep-learning-based methods in Discussion.

\section{Methods}
\subsection{Overview of DEN}
We propose a generic framework, called Deep Embedding Network, to accomplish nonlinear DR through DNNs. In short, DEN is an unsupervised learning method which uses DNN to optimize a dedicated loss function such that the output low-dimensional representation optimally preserves the desirable geometrical properties of the input high-dimensional data. Once trained, it forms a parametric mapping from input to output, able to embed any out-of-sample data. DEN is flexible in its design: it can make use of any modern modules of DNN, take in data of different forms (i.e. not only vectors but also tensors), and allow different loss functions.  

\subsection{Flexible Architecture: FCNN and CNN}
\begin{figure*}[!b]
\centering
\includegraphics[width=6in]{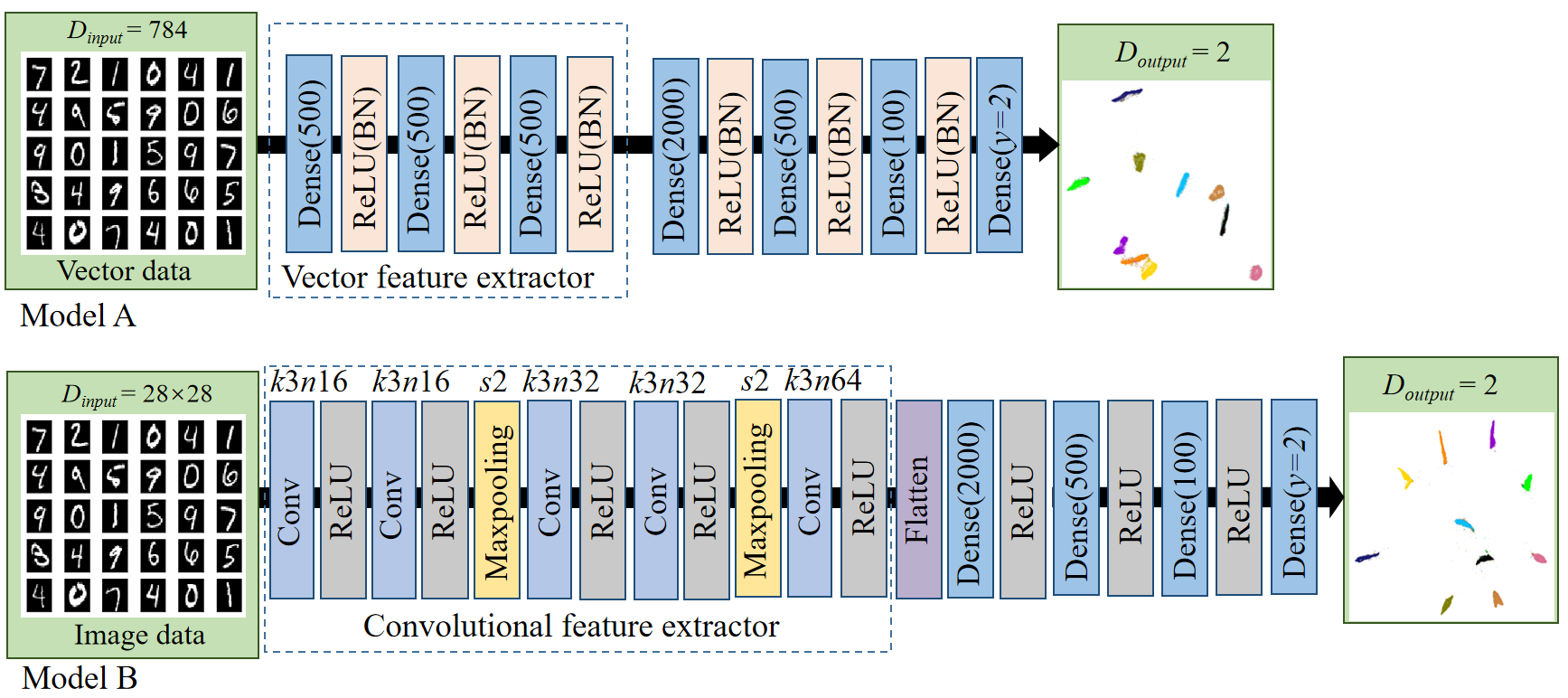}
\captionsetup{justification=centering}
\caption{Two exemplar architectures of the DEN model. Model A is a FCNN for vector input, and Model B is a CNN for image input. The CNN modules in Model B allow better extraction of features from images.}
\label{fig:1}
\end{figure*}

DEN allows very flexible design of the network architecture. Generally speaking, DEN can be divided into two parts: a feature extraction part and a dimensionality reduction part. In Fig. \ref{fig:1} we present two DEN architectures, named as Model A and Model B. Model A is a classical fully connected neural network (FCNN), while Model B includes convolutional neural network (CNN) modules for better extraction of features from images. 

Each network consists of two parts:
(1) Feature extraction: In this part, we aim to capture the high-level features from input data. In our experiments, we determined empirically the parameters of NN, and we noticed that the final performance is very robust to the setting of NN parameters. In Model A, the vector feature extractor contains 5 hidden layers, each of which followed by a ReLU and a batch normalization layer. In Model B, a standard design is followed: the first 2 convolutional layers (filter number=16) with ReLU and maxpooling capture shallow features, then 2 convolutional layers with 32 filters followed by ReLU and maxpooling operation are used to extract deeper features related to the task. 

(2) Dimension reduction: In this part, the dimensionality is gradually reduced, from 2000 to 500, then 100, and finally to the target dimension 2. We also followed a standard design in DNN: the first dense layer with 2000 neurons is used to lift the representations so that the capability to express data is increased. Then 2 dense layers with 500 and 100 neurons are used to reduce the dimension, as is often in NN design, i.e. a decreasing number of neurons at deeper layers. We note that the final DR results are not sensitive to the network hyperparameters.

\subsection{Flexible Loss: Deep t-SNE and Deep UMAP}
With the generic DEN architectures, we formulate dedicated loss functions to guide the training. DEN can take flexible loss functions, including that of t-SNE \citep{JMLR:v9:vandermaaten08a} and UMAP \citep{McInnes2018}, both in an information-theoretic sense. 

t-SNE \citep{JMLR:v9:vandermaaten08a} minimizes the Kullback-Leibler (KL) divergence between two estimated probability distributions, $P$ and $Q$, where $P$ is for $x$ in the high-dimensional space, and $Q$ for their representation $y$ in the low-dimensional space:
\begin{equation}
    L_\textrm{t-SNE} = \textrm{KL}(P||Q) = \sum_{i}\sum_{j\ne i}p_{ij}\log\frac{p_{ij}}{q_{ij}}
    \label{tsne-eq}
\end{equation}
where $p_{ij}$ and $q_{ij}$ are normalized pairwise similarities: $p_{ij}$ is the probability that a data point would choose the data point $x_j$ as its neighbor under the Gaussian distribution in the high-dimensional space, and $q_{ij}$ is the probability that a data point $y_i$ would choose the data point $y_j$ as its neighbor under the t-distribution in the low-dimensional space. $p_{ij}$ and $q_{ij}$ are calculated as follows:
\begin{alignat}{3}
    p_{j|i} & = \frac{\exp\left(\frac{{-\lVert x_i-x_j\rVert}^2}{2\sigma_i^2}\right)}{\sum_{k\ne i}\exp\left(\frac{{-\lVert x_i-x_k\rVert}^2}{2\sigma_i^2}\right)} \\
    p_{ij} & = \frac{p_{j|i}+p_{i|j}}{2N} \\
    q_{ji} & = \frac{\left(1+{\lVert y_i-y_j\rVert}^2\right)^{\frac{-\alpha -1}{2}}}{\sum_{k}\sum_{l\ne k}\left(1+{\lVert y_k-y_l\rVert^2}\right)^{\frac{-\alpha -1}{2}}}
\end{alignat}
where N is the number of data points, $\sigma_i$ is the Gaussian kernel at $x_i$, computed through a user defined perplexity \citep{JMLR:v9:vandermaaten08a}, and $\alpha$ is the degree of freedom of t-distribution. 

UMAP minimizes the fuzzy set cross entropy (CE) between two fuzzy membership functions \citep{McInnes2018}, $V$ and $W$, where $V$ is for $x$ in the high-dimensional space, and $W$ for their representation $y$ in the low-dimensional space. The loss is defined as:
\begin{align}
    L_\textrm{UMAP} & = \textrm{CE}(V||W) \\
    & = \sum_{i}\sum_{j\ne i}\left(v_{ij}\log\frac{v_{ij}}{w_{ij}}+(1-v_{ij})\log\frac{1-v_{ij}}{1-w_{ij}}\right)
    \label{umap-eq}
\end{align}

In the high-dimensional space, $v_{ij}$ is calculated as:
\begin{alignat}{2}
    v_{j|i} & = \exp\left(\frac{-d(x_i,x_j)+\rho_i}{\sigma_i}\right) \\
    v_{ij} & = (v_{j|i}+v_{i|j})-v_{j|i}v_{i|j}
\end{alignat}
where $d(x_i,x_j)$ is the distance between data point $x_i$ and $x_j$. $\rho_i$ is the minimum distance between $x_i$ and its neighbors to ensure connectivity of the map. $\sigma_i$ is derived as follows:
\begin{equation}
    \sum_{j=1}^k\exp\left(\frac{-\max(0,d(x_i,x_j)-\rho_i)}{\sigma_i}\right) = \log_2(k)
\end{equation}
where $k$ is the number of nearest neighbors when constructing the UMAP graph.

In the low-dimensional space, $w_{ij}$ is calculated as:
\begin{equation}
    w_{ij} = \left(1+a\lVert y_i-y_j\rVert_2^{2b}\right)^{-1}
\end{equation}
where $a$ and $b$ are user-defined values \citep{McInnes2018}.

We name the DEN trained with the t-SNE and UMAP loss \emph{deep t-SNE} and \emph{deep UMAP}, respectively. In the training process, mini-batches are used to reduce memory consumption for the expensive computation of $P$ and $V$. We first randomly shuffle the entire dataset, then divide the dataset into mini-batches of a fixed size. After that, the batches are trained one-by-one. When all batches of the entire dataset have been trained, one epoch is accomplished. The loss calculation is as define in (\ref{tsne-eq}) and (\ref{umap-eq}), and the two probability distributions $P$ and $Q$ for t-SNE (or $V$ and $W$ for UMAP) are calculated from data within mini-batches.

\subsection{Deep Recursive Embedding}
With introduction of the modern DNN components such as ReLU activation and batch normalization, we can largely improve the ease and efficiency of training, compared with the RBM-based ptSNE. However, we note that the upper limit of the embedding performance remains the same as that of the original t-SNE or UMAP, given that the loss function is identical to its original definition. To push the limit, we further present a recursive training strategy, called deep recursive embedding (DRE). 

DRE is proposed to make use of the latent representations extracted by DEN. In principle, the intermediate feature maps from different layers of DNN reflect different levels of abstraction of the original input. As discussed in \citep{Yosinski2015, Selvaraju2019}, the deeper layers of DNN provide more abstract representations specific to the learning task. In \citep{pmlr-v119-elthakeb20a}, the authors demonstrated that the deep layers of DNNs can extract a rich set of features, and leveraging these intermediate representations for knowledge distillation led to significant improvement of network efficiency. In \citep{Johnson2016}, the high-level intermediate feature maps extracted from a pretrained DNN are used to compute the perceptual loss to improve the image reconstruction performance. For our purpose, the high-level representations obtained by Deep t-SNE (or Deep UMAP) also contain relevant information related to the task of DR. Although unsupervised, these representations are optimized with the information-theoretic loss that strives to match high-dimensional and low-dimensional statistics. Such representations therefore serves as proper representation of the original data to progressively improve the performance of DR.

\begin{figure*}[!t]
\centering
\includegraphics[width=\linewidth]{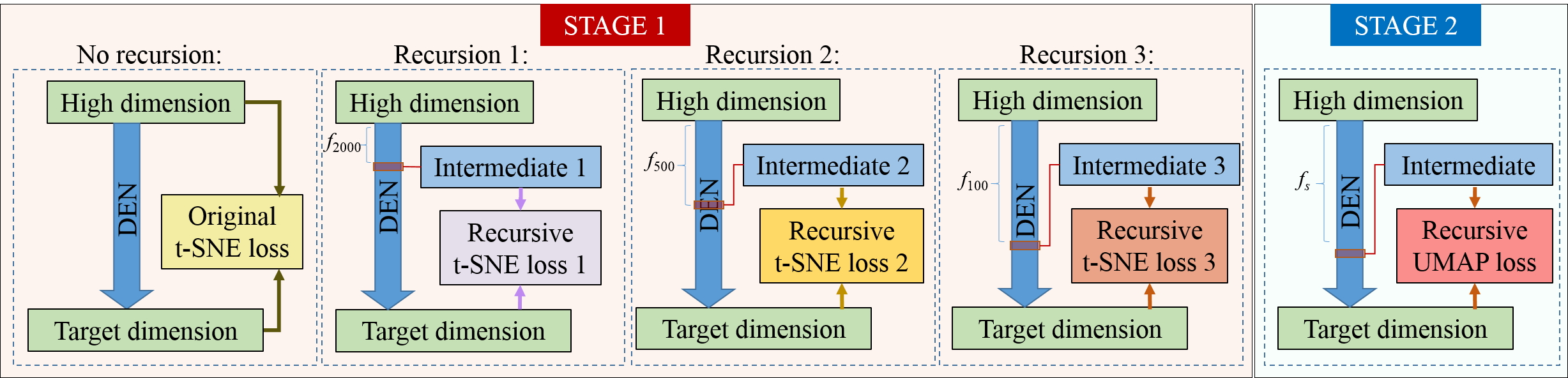}
\captionsetup{justification=centering}
\caption{The schematic of the proposed DRE method: Stage 1 shows the recursive training of DEN using the t-SNE loss, while Stage 2 further updates the network with the UMAP loss.}
\label{fig:2}
\end{figure*}

We propose to make use of the intermediate output of DEN for computation of $P$ or $V$. This process is therefore recursive, as illustrated in Fig. \ref{fig:2}. In each recursion step, we invoke intermediate outputs of feature extraction. Then, based on these extracted feature maps, we calculate a new probability distributions $\tilde{p}_{j|i}$ and $\tilde{v}_{j|i}$ for $\tilde{P}$ and $\tilde{V}$:
\begin{alignat}{2}
    \tilde{p}_{j|i} & = \frac{\exp\left(\frac{{-\lVert f(x_i)-f(x_j)\rVert}^2}{2\sigma_i^2}\right)}{\sum_{k\ne i}\exp\left(\frac{{-\lVert f(x_i)-f(x_k)\rVert}^2}{2\sigma_i^2}\right)} \\
    \tilde{v}_{j|i} & = \exp\left(\frac{-d\left(f(x_i),f(x_j)\right)-\rho_i}{\sigma_i}\right)
\end{alignat}
where f (·) is high-level feature extraction function realized by the feature extraction part of DEN. Based on this new probability distributions, we can define the recursive t-SNE loss and recursive UMAP loss:
\begin{alignat}{2}
    L_\textrm{t-SNE}^{'} & = \textrm{KL}(\tilde{P}||Q) \\
    L_\textrm{UMAP}^{'} & = \textrm{CE}(\tilde{V}||W)
\end{alignat}
At the beginning of recursive training, we calculate $P$ and $V$ once to estimate the loss in (\ref{tsne-eq}) and (\ref{umap-eq}). In principle, we can also continuously update $P$ and $V$ during the training. However, this turned to be time-consuming and our experiments showed that the improvement was minor.

\subsection{Multi-Stage Losses}
Our design also allows flexible combination of loss functions at different stage of recursion. In theory, UMAP pushes the dispersed points between different clusters to the nearest ones, driven by the second term in its loss function (\ref{umap-eq}). This results in a desirable visual effect promoting clean margins between classes (albeit not necessarily better classification performance or representation, as show in Results). We exemplify the flexible design of DEN by introducing a two-stage loss: first optimize DEN by the t-SNE loss, then refine the final visualization by the UMAP loss. The workflow is as follows: 

In stage 1, the network is trained with the original t-SNE loss $L_\mathrm{t-SNE}$ between $y$ and $x$; then, the network is recursively updated by minimizing the recursive t-SNE loss $L_\mathrm{t-SNE}^{'}$ between $y$ and the latent representations, generated by the output of Dense 2000, Dense 500 and Dense 100 layers, where the feature extractors are defined as $f_{2000} (\cdot)$, $f_{500} (\cdot)$, and $f_{100} (\cdot)$, respectively. This is shown as Recursion 1, Recursion 2 and Recursion 3 in Fig. \ref{fig:2} (STAGE 1). 

In stage 2, the DEN is further fine-tuned with the UMAP loss. This step is optional and for visualization purpose. This is shown in Fig. \ref{fig:2} (STAGE 2).

\section{Experiments}
\subsection{Datasets}

We experimented with 5 public datasets, covering different data types, sizes, and characteristics: (1) MNIST \cite{Lecun1998}, the handwritten digits dataset widely used to evaluate machine learning algorithms, which contains 60,000 training images and 10,000 testing images with size 28×28, in 10 classes from digit 0 to dig-it 9. (2) Fashion-MNIST \cite{Xiao2017}, the fashion product database by Zalando, developed in the same format as MNIST, containing 60,000 training images and 10,000 testing images of size 28×28, including 10 classes: T-shirt, trousers, pullover, dress, coat, sandal, shirt, sneaker, bag, and ankle boot. Fashion-MNIST is another benchmark database for machine learning algorithms, more challenging than MNIST. (3) RNA-Seq dataset includes 23,822 single-cell transcriptomes. The cells were isolated from the primary visual cortex (VISp) and anterior lateral motor cortex (ALM) of adult mouse. The cluster label is defined in \cite{Tasic2018}, which is used for visualizing different cell types. (4) IMDB dataset \cite{maas2011learning} includes 25,000 movie rating data used for sentiment analysis. All data are classified into positive and negative comments. Each comment was preprocessed to transform the textual sequences into a 500-dimensional word-vector. (5) InfiMNIST \cite{10.5555/2986916.2987026}: an open source method to generate an infinitely large database of handwritten digits based on MNIST. This dataset is used to test the scalability of DR algorithms. 

\subsection{Implementation}
Our environment is a Linux workstation with a NVIDIA Tesla V100 GPU, with 50GB system memory and 16GB GPU memory. The proposed DEN has a number of loss-related and network-related hyperparameters. In our experiments, we used Model A in Fig. \ref{fig:1} for vector input including RNA-Seq and IMDB, and Model B in Fig. \ref{fig:1} for image input including MNIST, Fashion-MNIST, and InfiMNIST. For the loss-related hyperparameters, we set the perplexity to 30 and the degree of freedom of t-distribution to 1. For the network-related hyperparameters, a mini-batch size of 2500 is chosen. Given one mini-batch of 2500 randomly selected data points from the input, $P_{ij}$ is computed to derive the gradient for updating the network parameters. Then the next mini-batch follows to update the network parameters further. The Adam optimizer was used for DNN training, with an initial learning rate of $10^{-3}$, $\beta_1$ (the exponential decay rate for the 1st moment estimate) of 0.9, $\beta_2$ (the exponential decay rate for the 2nd moment estimate) of 0.999 and an epsilon of $10^{-7}$. The training epochs of Deep t-SNE is set as 100. The DRE method has the first 100 epochs trained with the original t-SNE loss, and 50 epochs for each subsequent recursion in Stage 1. In Stage 2, an additional 100 epochs were trained with the UMAP loss.

\subsection{Evaluation Metrics}
We used 6 metrics to evaluate the DR performance \cite{Espadoto2021, Nonato2019}: (1) 1-nearest neighbor (1NN), (2) neighborhood hit, (3) trustworthiness, (4) continuity, (5) Shepard goodness, and (6) normalized stress. The definition of evaluation metrics are given as follows, where the dataset is denoted as $D=\{x_i\}$, $i=1,...,N$, $N$ is the number of sample points. 

(1) The 1-nearest neighbor (1NN) classification accuracy is a standard measurement as reported in \cite{JMLR:v9:vandermaaten08a, pmlr-v5-maaten09a}, which can be used to estimate the quality of clustering. With its absolute simplicity, the 1NN classification accuracy can be an indication of the goodness of clustering in the low-dimensional space.

(2) The neighborhood hit measures how well separable the data is in the low-dimensional space, which helps gauge if a technique is good for data exploration. The neighborhood hit is defined as:
\begin{equation}
    \sum_{i=1}^N\frac{j\in N_{i}^{(K)}:l_j=l_i}{KN},
\end{equation}
which indicates the proportion of $K$ neighbors $N_i^{(K)}$ of a point $i$ in the low-dimensional space who have the same label $l$ as point $i$ itself, averaged over all points in the low-dimensional space ($K=7$ as commonly used in literature).

(3) The trustworthiness measures the proportion of points in $D$ that are also close in its mapping, suggesting how much one can trust the local patterns in a projection. The trustworthiness is defined as:
\begin{equation}
    1-\frac{2}{NK(2n-3K-1)}\sum_{i=1}^N\sum_{j\in U_i^{(K)}}\left(r(i,j)-K\right),
\end{equation}
where $r(i,j)$ represents the rank of the low-dimensional data point $j$ according to the pairwise distances between the low-dimensional data points, and $U_i^{(K)}$ represents the set of points that are among the $K$ nearest neighbors in the low-dimensional space but not in the high-dimensional space. 

(4) The continuity measures the proportion of points in its mapping that are also close together in its original space, and is closely related to the missing neighbors of a projected point. The continuity is defined as:
\begin{equation}
    1-\frac{2}{NK(2n-3K-1)}\sum_{i=1}^N\sum_{j\in V_i^{(K)}}\left(\hat{r}(i,j)-K\right),
\end{equation}
where $\hat{r}(i,j)$ represents the rank of the high dimensional data point $j$ according to the pairwise distances between the low-dimensional data points, and $V_i^{(K)}$ represents the set of points that are among the $K$ nearest neighbors in the high-dimensional space but not in the low-dimensional space.

(5) The normalized stress measures the relative preservation of point-pairwise distances, which can be expressed as:
\begin{equation}
    \frac{\sum_{i,j}\big(\Delta^L(x_i,x_j)-\Delta^H(P(x_i),P(x_j))\big)^2}{\sum_{i,j}\Delta^L(x_i,x_j)},
\end{equation}
where $\Delta^L$ and $\Delta^H$ are distance metrics for data points in low- and high-dimensional space, respectively.

(6) The Shepard goodness measures the overall distance preservation by computing the Spearman rank correlation of Shepard diagram. The formula of generating a Shepard diagram is as follows:
\begin{equation}
    \mathrm{Scatterplot}\big(\lVert x_i-x_j\rVert,\lVert P(x_i)-P(x_j)\rVert\big), 1\le i\le N, i\ne j.
\end{equation}

\section{Results}
\subsection{Effect of Recursive Training}
\begin{figure*}[!t]
\centering
\includegraphics[width=6in]{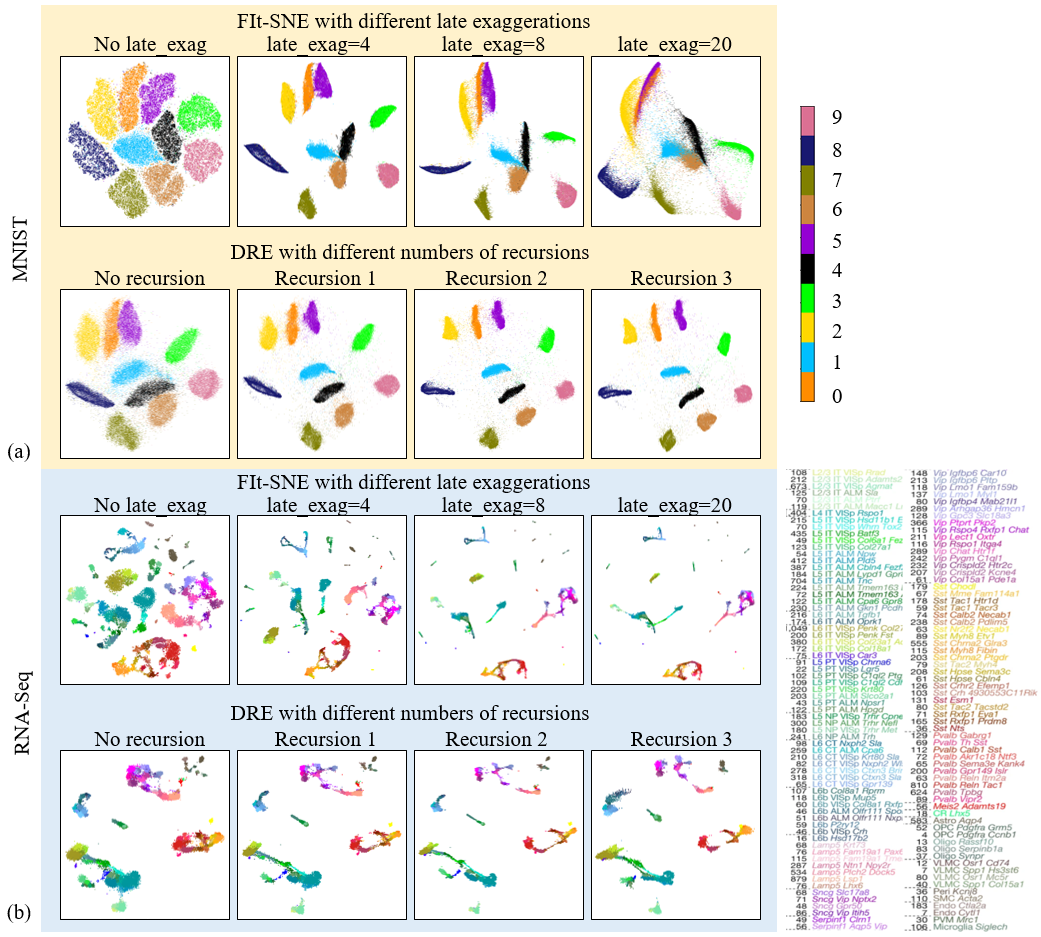}
\caption{Comparison of embedding performances of FIt-SNE with different late exaggeration settings and DRE with different recursions. (a) and (b) are results from the MNIST and RNA-Seq dataset, respectively. Different colors indicate different classes. Note that the color is for visualization, while the mapping was unsupervised in both FIt-SNE and DRE.}
\label{fig:3}
\end{figure*}

During recursive training, we observed that the clusters were gradually refined to better reflect the inter-cluster relationship, resulting in better separable clusters. This is reminiscent of the late exaggeration in FIt-SNE \cite{Linderman2019}, which also forces clusters to be tighter and more apart. For an empirical comparison between FIt-SNE and DRE, we implemented FIt-SNE with different late exaggeration settings, as well as DRE with different number of recursions. Fig. \ref{fig:3} shows a comparison between the two different strategies on the MNIST and RNA-Seq datasets. We observed that both late exaggeration and recursive training resulted in similar effect of map optimization by shortening the distance between data points within the same class while prolonging the distance between data points in different classes, both in an unsupervised manner. While FIt-SNE realizes this effect by increasing the repulsive forces explicitly during optimization, our method recalculates the distribution $P$ using an updated representation of the original high-dimensional data, obtained through DNN. The two strategies are not equivalent \emph{per se}, but both lead to a better separation of the inherent clusters in data. Closer observation shows that the DRE preserved a better visual balance between global and local structures. Within clusters we can better appreciate sub-structures in data, especially in the RNA-Seq data (lower panel b) with a natural hierarchical structure. 

\subsection{Comparison with other Embedding Methods}
We compared the proposed DEN methods with other reference methods on the MNIST, Fashion MNIST, RNA-Seq and IMDB datasets. For illustrations We included in total 7 methods for visual evaluation: (1) PCA, (2) AE, (3) t-SNE (Barnes-Hut implementation in Python), (4) UMAP, (5) DL Projection \cite{Espadoto2020} (trained by t-SNE), (6) DRE (with t-SNE loss), (7) DRE (with t-SNE and UMAP loss). Quantitative evaluation metrics are reported in Table \ref{table:1}, which compared a more extensive set of embedding methods. The comparison further includes RBM-based ptSNE \cite{pmlr-v5-maaten09a}, Fit-SNE \cite{Linderman2019}, DL Projection trained by UMAP \cite{Espadoto2020}, and Deep t-SNE (i.e. without recursion). 

Fig. 4 shows the results of MNIST and Fashion-MNIST, both in training and testing. For the t-SNE method (column c), no testing results was plotted as there is no learning mechanism. For MNIST, it can be observed that the PCA method (column a) could hardly differentiate the 10 classes, with poor performance in terms of 1NN error and neighborhood hit. The AE method (column b) performed better than PCA in terms of 1NN error, neighborhood hit, and trustworthiness, but the embedding remained visually poor. The Barnes-Hut t-SNE (column c) and UMAP (column d) are the two currently most popular visualization methods, which generated largely improved embeddings over PCA and AE. The DL projection method (column e) method reproduced the results of t-SNE (column c) both in training and testing. However, we noticed a change of cluster distribution (colunm c and e), due to the different optimization of t-SNE in separate runs. The last two columns show the embedding results from the proposed DRE method, with t-SNE loss (column f) and t-SNE + UMAP loss (column g), respectively. Both exhibited visually improved 2D embeddings. Even without color coding, we can still observe 10 distinct clusters, better separated compared with t-SNE or UMAP. Quantitatively, our proposed DRE method also showed superior 1NN and neighborhood hit value, comparable trustworthiness, continuity and normalized stress compared with those from the t-SNE and UMAP methods. Nonetheless, we noticed that the Shepard goodness was reduced, as the recursive training strategy exaggerated the inter-cluster distances, while Shepard goodness is a metric based on the original data distribution.

The Fashion-MNIST dataset is generally more challenging to cluster or classify than the MNIST dataset. We observed more distinct clusters (e.g. tops, bottoms and shoes) by our method than UMAP or t-SNE, however, the confusion of certain classes (e.g. shirt, pullover and coat) remained similar. 
\begin{figure*}[!t]
\centering
\includegraphics[width=\linewidth]{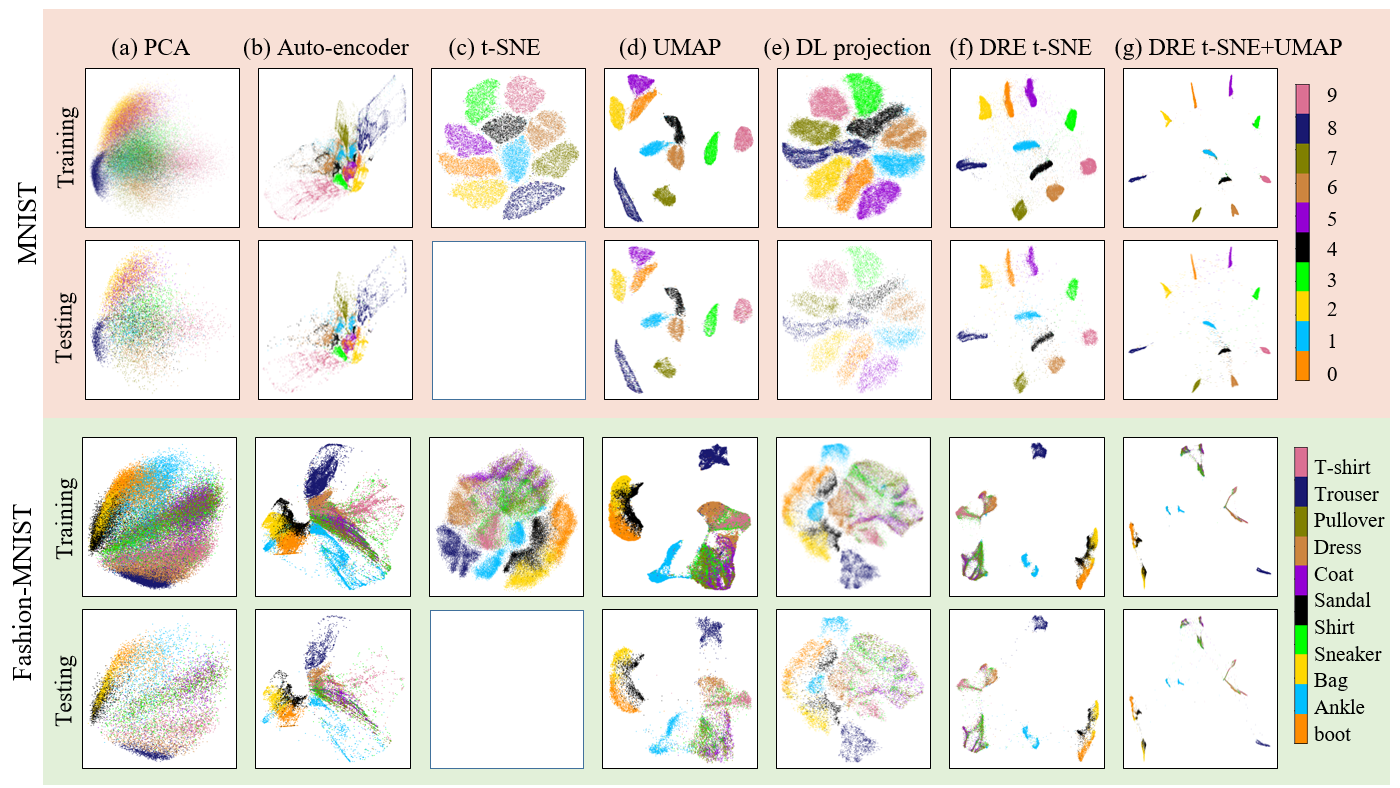}
\caption{Comparison of embedding performance on the MNIST and Fashion-MNIST datasets by 7 methods: (a) PCA, (b) AE, (c) t-SNE (Barnes-Hut implementation in Python), (d) UMAP, (e) DL Projection (trained by t-SNE), (f) DRE (with t-SNE loss), (g) DRE (with two-stage t-SNE + UMAP loss). The results of both training and testing datasets are shown.}
\label{fig:4}
\end{figure*}

\begin{figure*}[!t]
\centering
\includegraphics[width=\linewidth]{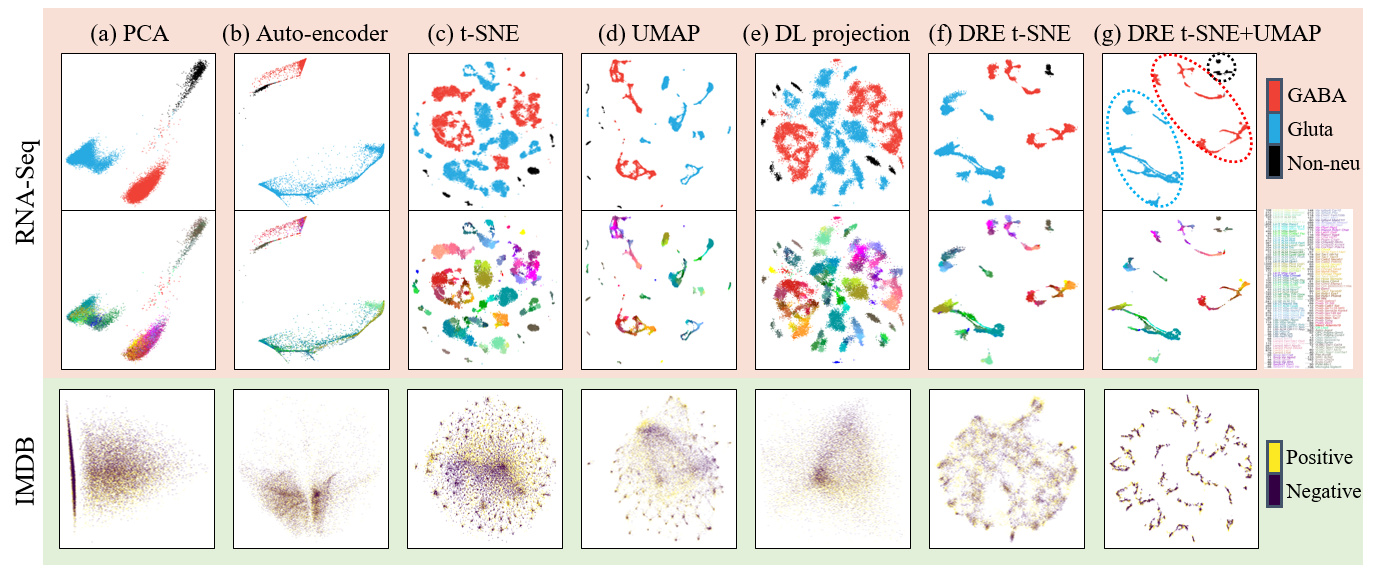}
\caption{Comparison of embedding performance on the RNA-Seq and IMDB datasets by 7 methods: (a) PCA, (b) AE, (c) t-SNE (Barnes-Hut implementation in Python), (d) UMAP, (e) DL Projection (trained by t-SNE), (f) DRE (with t-SNE loss), (g) DRE (with two-stage t-SNE + UMAP loss). For readability of figures only the results of training datasets are shown. For RNA-seq, we showed two color schemes to highlight the hierarchy of global and local data structure.}
\label{fig:5}
\end{figure*}

The results of RNA-Seq are shown in the first and second rows of Fig. \ref{fig:5}, in two color schemes reflecting the hierarchy of global and local data structure. The RNA-Seq can be divided into three major types: glutamatergic excitatory neurons (GABA), GABAergic inhibitory neurons (Gluta) and non-neuronal cells (Non-neu), color-coded in red, blue and black in the first row. It can be observed that the PCA method could differentiate the three major types of cells very well, presenting the global structure in data. In each major cell type, there are also a large number of subtypes, in total 133 (details can be find in \cite{Tasic2018}). The second row of Fig. \ref{fig:5} color-codes the cell subtypes in a refined color scale. As the local data structures are nonlinear, PCA can no longer differentiate these subtypes, in contrast to nonlinear methods such as t-SNE and UMAP, which exhibited clustering according to cell subtype. However, we observed a loss of global structure in the results of t-SNE and UMAP, with the global distribution of three major types no longer consistent, i.e. scattered and mingled. The proposal DRE method (column f and g) showed a better balance between global and local data structures. The three major types are close as highlighted by the dotted ellipses in Fig. \ref{fig:5} column g, whereas the subtypes can also be appreciated. 

The results of IMDB dataset is shown in the third row of Fig. \ref{fig:5}. The original data are only classified into positive and negative comments, however, the actual sentiments are much more complicated and subtle. We observed that all embedding methods mixed up the two categories of data, but nonlinear methods (column c-g) showed sub-clusters, which might reflect subtle semantics in comments. In Table \ref{table:1}, we note that the two DRE methods resulted in better quantitative metrics compared to other methods, with the highest neighborhood hit, trustworthiness and Shepard goodness, and lowest normalized stress. However, further evaluation demands more detailed labels of sub-clusters.

\Heatset{min=0.079,   
        mid=0.36,
        max=1,   
        max colour=SpringGreen4, 
        mid colour=Khaki1, 
        min colour=OrangeRed1,      
        Min colour=OrangeRed1, 
        Max colour=Green1   
}
\begin{table*}[!t]
\centering
\caption{Comparison of the embedding performance on MNIST, Fashion-MNIST, IMDB and RNA-Seq data, by different methods. $M_{NH}$, $M_{t}$, $M_C$, $M_{\sigma}$, $M_S$ indicate the values of neighborhood hit, trustworthiness, continuity, Shepard goodness, and normalized stress, respectively.}
\resizebox{\linewidth}{!}{%
\begin{tabular}{ccHHHHHHHHHHH} 
 \hline\hline
 \multicolumn2c{Experiments} & \multicolumn1c{PCA} & \multicolumn1c{Auto-} & \multicolumn1c{t-SNE} & \multicolumn1c{FIt-SNE} & \multicolumn1c{UMAP} & \multicolumn1c{RBM} & \multicolumn1c{DL projection} & \multicolumn1c{DL projection} & \multicolumn1c{Deep} & \multicolumn1c{DRE} & \multicolumn1c{DRE} \\
 & & & \multicolumn1c{encoder} & \multicolumn1c{(Barnes-Hut)} & & & \multicolumn1c{ptSNE} & \multicolumn1c{(t-SNE)} & \multicolumn1c{(UMAP)} & \multicolumn1c{t-SNE} & \multicolumn1c{t-SNE} & \multicolumn1c{t-SNE + UMAP} \\
 \hline
 & $M_{NH}$ & 0.47 & 0.80 & 0.93 & 0.92 & 0.92 & 0.81 & 0.90 & 0.95 & 0.93 & 0.96 & 0.97 \\
 & $M_{t}$ & 0.74 & 0.94 & 0.98 & 0.95 & 0.96 & 0.92 & 0.92 & 0.93 & 0.94 & 0.94 & 0.92 \\
 \rotatebox{90}{\makebox[0pt]{MNIST}}
 & $M_{C}$ & 0.94 & 0.96 & 0.98 & 0.98 & 0.97 & 0.97 & 0.97 & 0.97 & 0.97 & 0.97 & 0.96 \\
 & $1-M_{\sigma}$ & 0.42 & 0.43 & 0.54 & 0.54 & 0.50 & 0.24 & 0.53 & 0.54 & 0.53 & 0.54 & 0.52 \\
 & $M_{S}$ & 0.50 & 0.43 & 0.43 & 0.39 & 0.37 & 0.53 & 0.32 & 0.31 & 0.34 & 0.29 & 0.29 \\
 \hline
 & $M_{NH}$ & 0.53 & 0.71 & 0.77 & 0.77 & 0.73 & 0.70 & 0.71 & 0.73 & 0.73 & 0.76 & 0.77 \\
 & $M_{t}$ & 0.91 & 0.97 & 0.99 & 0.99 & 0.98 & 0.97 & 0.96 & 0.96 & 0.97 & 0.97 & 0.97 \\
 \rotatebox{90}{\makebox[0pt]{Fashion-MNIST}}
 & $M_{C}$ & 0.98 & 0.98 & 0.99 & 0.99 & 0.99 & 0.99 & 0.98 & 0.99 & 0.99 & 0.98 & 0.98 \\
 & $1-M_{\sigma}$ & 0.65 & 0.50 & 0.61 & 0.64 & 0.56 & 0.63 & 0.61 & 0.65 & 0.62 & 0.60 & 0.62 \\
 & $M_{S}$ & 0.88 & 0.72 & 0.58 & 0.65 & 0.58 & 0.76 & 0.50 & 0.63 & 0.63 & 0.60 & 0.59 \\
 \hline
 & $M_{NH}$ & 0.59 & 0.59 & 0.65 & 0.62 & 0.65 & 0.59 & 0.61 & 0.62 & 0.59 & 0.72 & 0.76 \\
 & $M_{t}$ & 0.69 & 0.62 & 0.89 & 0.72 & 0.79 & 0.68 & 0.63 & 0.67 & 0.73 & 0.74 & 0.74 \\
 \rotatebox{90}{\makebox[0pt]{IMDB}}
 & $M_{C}$ & 0.66 & 0.63 & 0.84 & 0.80 & 0.85 & 0.70 & 0.75 & 0.76 & 0.79 & 0.78 &0.79 \\
 & $1-M_{\sigma}$ & 0.26 & 0.23 & 0.34 & 0.38 & 0.32 & 0.23 & 0.24 & 0.28 & 0.37 & 0.44 & 0.45 \\
 & $M_{S}$ & 0.16 & 0.08 & 0.29 & 0.27 & 0.26 & 0.37 & 0.34 & 0.29 & 0.39 & 0.38 & 0.36 \\
 \hline
 & $M_{NH}$ & 0.91 & 0.27 & 0.99 & 1.00 & 1.00 & 0.96 & 0.98 & 0.99 & 1.00 & 1.00 & 1.00 \\
 & $M_{t}$ & 0.85 & 0.82 & 1.00 & 0.99 & 0.99 & 0.98 & 0.98 & 0.98 & 0.99 & 0.98 & 0.99 \\
 \rotatebox{90}{\makebox[0pt]{RNA-Seq}}
 & $M_{C}$ & 0.93 & 0.91 & 1.00 & 1.00 & 0.99 & 0.99 & 0.99 & 0.98 & 0.99 & 0.99 & 0.99 \\
 & $1-M_{\sigma}$ & 0.53 & 0.51 & 0.59 & 0.61 & 0.61 & 0.56 & 0.55 & 0.61 & 0.62 & 0.61 & 0.60 \\
 & $M_{S}$ & 0.88 & 0.44 & 0.40 & 0.57 & 0.55 & 0.79 & 0.30 & 0.56 & 0.64 & 0.62 & 0.62 \\
 \hline\hline
\end{tabular}
}
\label{table:1}
\end{table*}

\subsection{Scalability and Generalization}
To evaluate the scalability of the embedding methods, we generated huge training and testing datasets from InfiMNIST \cite{10.5555/2986916.2987026}, and evaluated if the proposed DRE could scale to such extremely large dataset. We trained DRE with 0.05 million, 0.1 million and 0.3 million InfiMNIST data, and embedded an independently generated 1 million InfiMNIST testing dataset. Then we measured the 1NN classification accuracy on the 2D embedding, as reported in Table \ref{table:2}. The proposed DRE method out-performed all other embedding methods, with a 93.52\%, 95.58\% and 97.17\% testing accuracy when trained on 0.05 million, 0.1 million and 0.3 million data, respectively. Fig. 6 shows the embedding results of the 1 million testing data (trained on 0.1 million data), for 6 methods in comparison: (a) PCA, (b) AE, (c) UMAP, (d) DL projection trained with t-SNE, (e) DRE with t-SNE loss, and (f) DRE with t-SNE + UMAP loss, respectively. The embedding of both the training and testing datasets are shown. On the testing results, we zoomed in the local neighborhood at the inside and border of class 9 to inspect embedding details. As the learned embedding was extrapolated from 0.1 million training data onto 1.0 million testing data, we observed scattering of classes at the border. Overall, the proposed DRE methods resulted in the cleanest neighborhood compared to other embedding methods, which explains the high classification accuracy in Table \ref{table:2}.
\begin{figure*}[!t]
\centering
\includegraphics[width=6.5in]{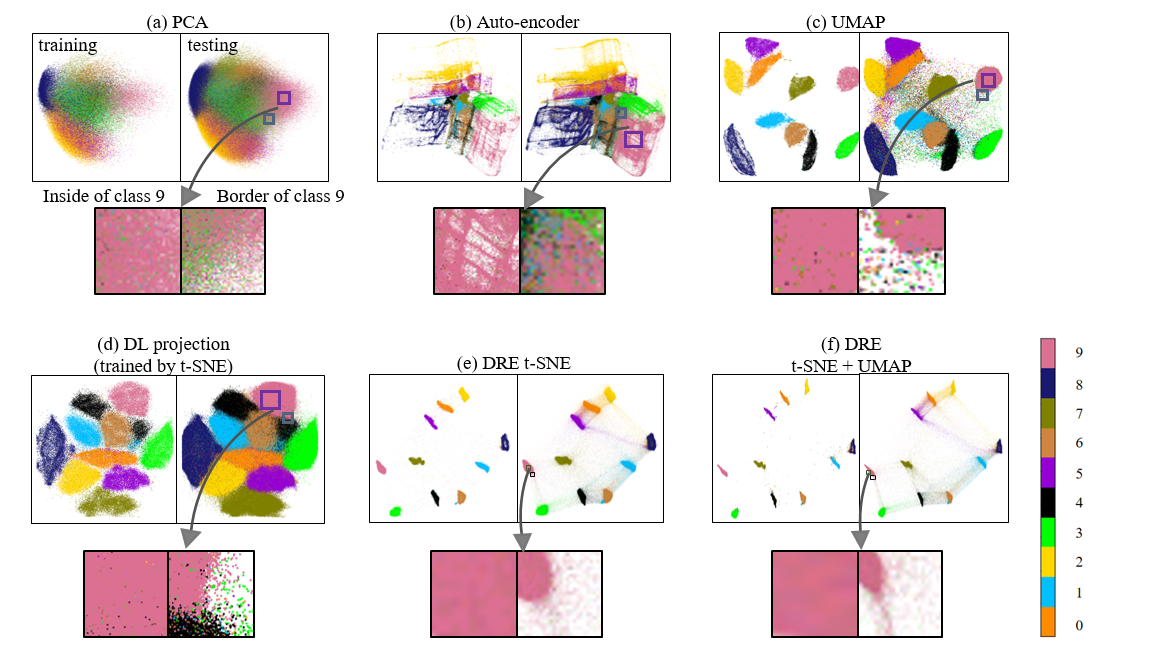}
\caption{Embedding 1 million InfiMNIST data by learning from 0.1 million data. The embeddings of both training and testing data are shown for (a) PCA, (b) AE, (c) UMAP, (d) DL projection trained with t-SNE, (e) DRE with t-SNE loss, and (f) DRE with t-SNE + UMAP loss. Local neighborhoods inside and at the border of class 9 are zoomed in for a closer look of embedding quality.}
\label{fig:6}
\end{figure*}

In a second experiment, we used a training set of 10K data points to train the embedding methods, and tested them on an increasing testing set of 10K, 30K, and 60K. The 1NN classification accuracy is reported in Table \ref{table:3}. It can be seen that all embedding methods showed stable generalization, suggesting the manifold of MNIST can be adequately learned and represented in a dimensionality as low as 2. In both Table \ref{table:2} and Table \ref{table:3}, we observed that the two DRE methods had the best quantitative performance, followed by deep t-SNE and the DL projection method. The results can be intuitively explained as follows: both deep t-SNE and DL projection methods are approaching the limit of t-SNE, while the DRE methods are pushing the limit further by recursive training. Comparing the two DRE methods using t-SNE and composite t-SNE+UMAP losses, however, we noticed little difference in 1NN classification performance or other quantitative measures, although the latter produces visually better separation and clearer margin.

\Heatset{min=35,   
        mid=85,
        max=97.2,   
        max colour=SpringGreen4, 
        mid colour=Khaki1, 
        min colour=OrangeRed1,      
        Min colour=OrangeRed1, 
        Max colour=Green1   
}
\begin{table*}[!t]
\centering
\caption{Comparison of the 1-nearest neighbor classification accuracy (\%) of the 1 million InfiMNIST data, embedded by learning from training data of different sizes: 0.05, 0.1, and 0.3 Million.}
\resizebox{0.85\linewidth}{!}{%
\begin{tabular}{cHHHHHHHHH} 
 \hline\hline
 \multicolumn1c{Training} & \multicolumn1c{PCA} & \multicolumn1c{Auto-} & \multicolumn1c{UMAP} & \multicolumn1c{RBM} & \multicolumn1c{DL projection} & \multicolumn1c{DL projection} & \multicolumn1c{Deep} & \multicolumn1c{DRE} & \multicolumn1c{DRE} \\
 \multicolumn1c{data size} & & \multicolumn1c{encoder} & & \multicolumn1c{ptSNE} & \multicolumn1c{(t-SNE)} & \multicolumn1c{(UMAP)} & \multicolumn1c{t-SNE} & \multicolumn1c{t-SNE} & \multicolumn1c{t-SNE + UMAP} \\
 \hline
 0.05M & 35.80 & 64.14 & 87.00 & 64.84 & 83.68 & 86.52 & 89.96 & 93.33 & 93.52 \\
 0.1M & 39.14 & 68.35 & 90.25 & 71.14 & 87.61 & 90.23 & 91.64 & 95.48 & 95.58 \\
 0.3M & 52.36 & 78.50 & 93.60 & 82.38 & 90.39 & 93.07 & 94.76 & 97.14 & 97.17 \\
 \hline\hline
\end{tabular}
}
\label{table:2}
\end{table*}

\Heatset{min=36,   
        mid=85,
        max=90.7,   
        max colour=SpringGreen4, 
        mid colour=Khaki1, 
        min colour=OrangeRed1,      
        Min colour=OrangeRed1, 
        Max colour=Green1   
}
\begin{table*}[!t]
\centering
\caption{Comparison of the 1-nearest neighbor classification accuracy (\%) on InfiMNIST testing datasets of different sizes: 10K, 30K, and 60K. The training data size is 10K.}
\resizebox{0.85\linewidth}{!}{%
\begin{tabular}{cHHHHHHHHHH}
 \hline\hline
 \multicolumn1c{Testing} & \multicolumn1c{PCA} & \multicolumn1c{Auto-} & \multicolumn1c{UMAP} & \multicolumn1c{RBM} & \multicolumn1c{DL projection} & \multicolumn1c{DL projection} & \multicolumn1c{Deep} & \multicolumn1c{DRE} & \multicolumn1c{DRE} \\
 \multicolumn1c{data size} & & \multicolumn1c{encoder} & & \multicolumn1c{ptSNE} & \multicolumn1c{(t-SNE)} & \multicolumn1c{(UMAP)} & \multicolumn1c{t-SNE} & \multicolumn1c{t-SNE} & \multicolumn1c{t-SNE + UMAP} \\
 \hline
 10K & 37.32 & 67.43 & 78.98 & 68.58 & 82.31 & 85.08 & 87.09 & 89.83 & 90.07 \\
 30K & 37.85 & 66.89 & 87.76 & 67.79 & 81.66 & 84.45 & 87.01  & 90.09 & 90.08 \\
 60K & 38.10 & 67.47 & 88.01 & 68.47 & 82.06 & 84.75 & 87.26 & 90.49 & 90.67 \\
 \hline\hline
\end{tabular}
}
\label{table:3}
\end{table*}

\subsection{Runtime}
We tested the runtime of a large range of existing embedding methods in literature, categorized into parametric and non-parametric DR methods. The runtime are reported in Table \ref{table:4}. The PCA, Eigenmaps and BH t-SNE methods were implemented single threaded on CPU. UMAP used numba's parallel implementation to do multithreaded processing with multiple cores on CPU. The DL projection methods adopted the MulticoreTSNE with 8 NVIDIA Tesla V100 GPU to parallelly calculate the network references, and fitted the model using 1 GPU. The t-SNE and FIt-SNE method utilized the multithreading and C/C++ compiler on CPU. The proposed methods single threaded all computations, and used 1 GPU to train the network. As shown in Table \ref{table:4}, the runtime of the proposed Deep t-SNE, DR t-SNE and DRE increased gradually with more recursions. Compared with non-parametric methods, the DRE method needed less runtime than the publicly available Open t-SNE \cite{Policar2019}, but more runtime than FIt-SNE \cite{Linderman2019}. Among the parametric methods, the proposed methods were faster than the RBM-based ptSNE. Although our methods are slower than AE and UMAP during training, they are very fast when applied on new data points (second row ``test").

\begin{table*}[!t]
\centering
\caption{Comparison of the runtime of different embedding methods (MNIST: 60,000 training and 10,000 testing)}
\resizebox{\linewidth}{!}{%
\begin{tabular}{c c c c c c c c c c c | c c c c} 
 \hline\hline
 \multicolumn2c{Type} & \multicolumn9{c|}{Parametric DR methods} & \multicolumn4{c}{Non-parametric DR Methods} \\
 \hline
    \multicolumn2c{Method} & \multicolumn1c{PCA} & \multicolumn1c{Auto-} & \multicolumn1c{RBM} & \multicolumn1c{UMAP} & \multicolumn1c{DL projection} & \multicolumn1c{DL projection} & \multicolumn1c{Deep t-} & \multicolumn1c{DR t-} & \multicolumn1{c|}{DRE} &  \multicolumn1c{Eigenmaps} & \multicolumn1c{BH t-} & \multicolumn1c{Open t-} & \multicolumn1c{FIt-}\\
    & & & \multicolumn1c{encoder} & \multicolumn1c{ptSNE} & & \multicolumn1c{(t-SNE)} & \multicolumn1c{(UMAP)} & \multicolumn1c{SNE} & \multicolumn1c{SNE} & & & \multicolumn1c{SNE} &  \multicolumn1c{SNE} & \multicolumn1c{SNE}\\
 \hline
 Runtime & Train & 0.029 & 31.78 & 7455.39 & 83.88 & 2122.01 & 1085.72 & 250.81 & 339.51 & 759.88 & 42026.70 & 5730.79 & 1149.47 & 304.17 \\
 (second) & Test & 0.67 & 0.46 & 0.49 & 13.68 & 0.22 & 0.40 & 0.84 & 0.82 & 0.73 & \diagbox[dir=SW]{}{} & \diagbox[dir=SW]{}{} & \diagbox[dir=SW]{}{} & \diagbox[dir=SW]{}{}\\
 \hline\hline
\end{tabular}
}
\label{table:4}
\end{table*}

\section{Discussion}
In this work, we propose to integrate modern DNNs and mathematics principles to embed high-dimensional data, and we name this framework ``deep embedding network'' - DEN. DEN is a generic framework, which can be designed in a flexible manner in terms of NN architectures, training strategies, and loss functions. Based on DEN, we showed how to design Deep t-SNE and Deep UMAP, and further introduced a recursive training strategy to boost the embedding performance over the original t-SNE or UMAP. We exemplified the flexibility of DEN by the recursive training with two-stage loss functions combining the popular t-SNE and UMAP. We tested the proposed DRE on a variety of public datasets, and evaluated its performance both visually and quantitatively, against a comprehensive set of classical and state-of-the-art embedding methods. 

While inspired by t-SNE, the proposed DRE has made the following improvement: (1) the DRE method is parametric and can easily embed new, out-of-sample data; (2) the proposed recursive training strategy can boost the embedding performance over the original t-SNE; (3) empirical experiments show that the DRE method can better preserve the global data structure simultaneously; (4) it can integrate any mathematics rules beyond t-SNE. 

Both DL projection method and the proposed DRE method can map large, out-of-sample datasets. This is an inherent advantage of learning-based methods. However, the training mechanisms are very different: DL Projection solves the DR problem by learning from the results of t-SNE or UMAP, i.e. the network is taught to generate an output similar to that of the reference methods. In contrast, the proposed DRE method provides the network a principled loss function, and then let the network explore the embedding itself.

There are a number of hyperparameters in the proposed method. The mini-batch used in DNN is critical for scalability and has an influence on the embedding results: a small mini-batch cannot fully sample the data distribution, while a big mini-batch demands too much memory as the square matrix of $P$ is dependent on mini-batch size. We empirically selected 2500, but it can also be set larger if memory allows. Other network settings only had minor influence on the final performance according to our experiments, given the capability of DNNs to optimize highly complex loss functions. Another hyperparameter is the number of the recursions, for which we can use a simple rule of thumb: if we are keen in visualizing the global clustering of data, we can increase the number of recursions (e.g. 3) and obtain better separated clusters in a global view; otherwise 1 or 2 recursions are sufficient as our experiments indicated. We argue that the improved global structure preservation by DRE may arise from the use of latent representations, which better capture the global structure in data (for example, PCA is a linear latent representation). Meanwhile, the local structures are captured by DRE as it is initiated from the t-SNE embedding rule, which emphasizes local neighborhood. 

There are a few practical weaknesses of the proposed methods. First, our method obtains the parametric mapping by learning from a dataset, so it is more suitable for big data datasets than for small datasets. Second, the proposed methods still need more runtime than PCA, AE and UMAP. Nevertheless, we expect that the runtime can be reduced by multi-threaded implementation.

\section{Conclusion}
In this paper, we introduced a generic DEN frame-work, which is flexible in its NN architectures, training strategies, and loss functions. Based on DEN, we proposed a novel DRE strategy to further boost the embedding performance. The proposed framework can combine modern DNNs and any effective embedding rules such as those from t-SNE and UMAP. The proposed DRE method can map out-of-sample data and scale to extremely large datasets. Experiments on a range of datasets demonstrated its improved embedding performance in terms of local and global structure preservation, compared with other state-of-the-art embedding methods.

\bibliographystyle{unsrtnat}
\bibliography{DRE}






\end{document}